\documentclass{article}

\usepackage[preprint]{neurips_2022}

\usepackage[utf8]{inputenc} 
\usepackage[T1]{fontenc}    
\usepackage{hyperref}       
\usepackage{url}            
\usepackage{booktabs}       
\usepackage{amsfonts}       
\usepackage{nicefrac}       
\usepackage{microtype}      
\usepackage{xcolor}         

\usepackage{algorithm}
\usepackage[noend]{algpseudocode}
\usepackage[font=small,labelfont=bf]{caption}
\usepackage{booktabs}
\usepackage{makecell}
\usepackage{amsmath}
\usepackage{bm}
\usepackage{graphicx}
\usepackage{amssymb}

\newcommand{\exalgo}{{\cal{A}}}

\newcommand{\lexp}{{\mathit{e}}}
\newcommand{\Dorig}{{\cal{D}}}

\newcommand{\buffer}{{\cal{M}}}

\newcommand{\loss}{{\cal{L}}}

\def\vx{{\bm{x}}}

\title{Architect, Regularize and Replay (ARR): a Flexible Hybrid Approach for Continual Learning}

\author{%
  Vincenzo Lomonaco\\
  Department of Computer Science\\
  University of Pisa\\
  \texttt{vincenzo.lomonaco@unipi.it} \\
   \And
   Lorenzo Pellegrini \\
   Department of Computer Science and Engineering \\
   University of Bologna \\
   \texttt{l.pellegrini@unibo.it} \\
   \AND
   Gabriele Graffieti \\
   Department of Computer Science and Engineering \\
   University of Bologna \\
   \texttt{gabriele.graffieti@unibo.it} \\
   \And
   Davide Maltoni \\
   Department of Computer Science and Engineering \\
   University of Bologna \\
   \texttt{davide.maltoni@unibo.it} \\
}

\begin{document}

\maketitle

\begin{abstract}
In recent years we have witnessed a renewed interest in machine learning methodologies, especially for deep representation learning, that could overcome basic i.i.d. assumptions and tackle non-stationary environments subject to various distributional shifts or sample selection biases. Within this context, several computational approaches based on architectural priors, regularizers and replay policies have been proposed with different degrees of success depending on the specific scenario in which they were developed and assessed. However, designing comprehensive hybrid solutions that can flexibly and generally be applied with tunable efficiency-effectiveness trade-offs still seems a distant goal. In this paper, we propose \emph{Architect, Regularize and Replay} (ARR), an hybrid generalization of the renowned AR1 algorithm and its variants, that can achieve state-of-the-art results in classic scenarios (e.g. class-incremental learning) but also generalize to arbitrary data streams generated from real-world datasets such as CIFAR-100, CORe50 and ImageNet-1000.
\end{abstract}

\section{Introduction}
Continual Machine Learning is a challenging research problem with profound scientific and engineering implications \citep{lomonaco2018thesis}. On one hand, it undermines the foundations of classic machine learning systems relying on \emph{iid} assumptions, on the other hand, it offers a path towards efficient and scalable human-centered AI systems that can learn and think like humans, swiftly adapting to the ever-changing nature of the external world. However, despite the recent surge of interest from the machine learning and deep learning communities on the topic and the prolific scientific activity of the last few years, this vision is far from being reached. 

While most continual learning algorithms significantly reduce the impact of catastrophic forgetting on specific scenarios, it is difficult to generalize those results to settings in which they have not been specifically designed to operate (lack of robustness and generality). Moreover, they are mostly focused on vertical and exclusive approaches to continual learning based on regularization, replay or architectural changes of the underlying prediction model.

In this paper, we summarize the effort made in the formulation of hybrid strategies for Continual Learning that can be more robust, generally applicable and effective in real-world application contexts. In particular, we will focus on the definition of the ``\emph{Architect}, \emph{Regularize} and \emph{Replay}" (ARR) method: a general reformulation and generalization of the renowned AR1 algorithm \citep{maltoni2019} with all its variants \citep{lomonaco2019continual, pellegrini2019latent}, and, arguably, one of the first hybrid continual learning methods proposed \citep{parisi2020online} (Sec. \ref{sec:arr}). 

Through a number of experiments on state-of-the-art benchmarks such as \texttt{CIFAR-100}, \texttt{CORe50} and \texttt{ImageNet-1000}, we show the efficiency and effectiveness of the proposed approach with respect to other existing state-of-the-art methods (Sec \ref{sec:eval}). Then, we discuss tunable parameters to easily control the effectiveness-efficiency trade-off such as the selection of the \emph{latent replay layer} (Sec. \ref{subsec:replay_layer_selection}) and the \emph{replay memory size} (Sec. \ref{subsec:replay_mem_size_selection}). Finally, we discuss current \texttt{ARR} implementation porting in Avalanche \citep{lomonaco2021avalanche} (Sec \ref{sec:avalanche}).

\section{Background and Problem Formulation}

\emph{Continual Learning} (CL) is mostly concerned with the concept of learning from a stream of ephemeral non-stationary data that can be processed in separate computational steps and cannot be revisited if not explicitly memorized. In an agnostic continual learning scenario data arrives in a streaming fashion as a (possibly infinite) sequence $S$ of, what we call, \emph{learning experiences} $e$, so that $S = \lexp_1, \hdots, \lexp_n$. For simplicity, we assume a supervised classification problem, where each experience $e_i$ consists of a batch of samples $\Dorig^i$, where each sample is a tuple $\langle x^i_k, y^i_k\rangle$ of input and target data, respectively, and the labels $y^i_k$ are from the set ${\cal{Y}}^i$, which is a subset of the entire universe of classes ${\cal{Y}}$.
However, we note this formulation is very easy to generalize to different CL problems. Usually $\Dorig^i$ is split into a separate train set $\Dorig^i_{train}$ and test set $\Dorig^i_{test}$. A continual learning algorithm~ $\exalgo^{CL}$ is a function with the following signature \citep{lesort:hal-02381343,carta2021ex}:
\begin{equation}
    \exalgo^{CL}:\ \langle f^{CL}_{i-1}, \Dorig^i_{train}, \buffer_{i-1}, t_i\rangle\ \rightarrow\ \langle f^{CL}_i, \buffer_{i}\rangle
\end{equation}
where $f^{CL}_i$ is the model learned after training on experience $\lexp_i$, $\buffer_i$ a buffer of past knowledge (can be also void), such as previous samples or activations, stored from the previous experiences and usually of fixed size. The term $t_i$ is a task label that may be used to identify the correct data distribution (or \emph{task}). All the experiments in this paper assume the most challenging scenario of $t_i$ being unavailable. 
Usually, CL algorithms are limited in the amount of resources that they can use and they should be designed to scale up to a large number of training experiences without increasing their memory / computational overheads over time. 
The objective of a CL algorithm is to minimize the loss $\loss_{S}$ over the entire stream of data $S$, composed of $n$ distinct experiences:
\begin{align}
    \loss_{S}(f^{CL}_n, n) = \frac{1}{\sum\limits_{i=1}^n |\Dorig_{test}^i|} \sum_{i=1}^n \label{eq:cl_objective} \loss_{exp}(f^{CL}_n, \Dorig^i_{test}) \\
    \loss_{exp}(f^{CL}_n, \Dorig^i_{test}) = \sum_{j=1}^{|\Dorig_{test}^i|} \loss(f^{CL}_n(\vx^i_j), y^i_j),
\end{align}
where the loss $\loss(f^{CL}_n(\vx), y)$ is computed on a single sample $\langle\vx, y\rangle$, such as cross-entropy in classification problems. Hence, the main assumption in this formulation is that all the concepts encountered over time are still relevant (the drift is only \emph{virtual}) and there's no conflicting evidence. This is quite a common assumption for the deep continual learning literature which is more concerned with building robust and general representations over time rather than building systems that can quickly adapt to changing circumstances. 

\section{Towards Hybrid Continual Learning Approaches}

\begin{figure}[t]
  \centering
  \includegraphics[width=0.55\textwidth]{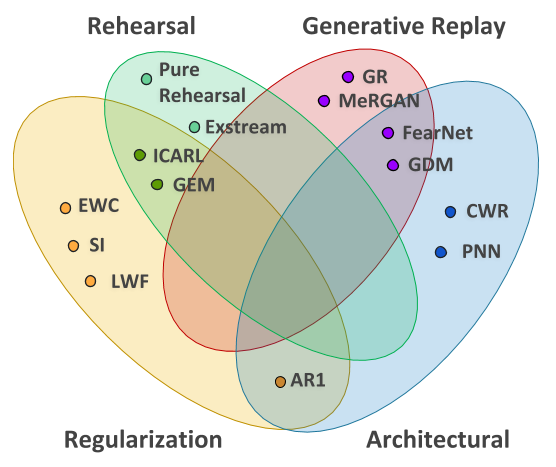}
  \caption{Venn diagram of some of the most popular CL strategies: CWR \citep{Lomonaco17}, PNN \citep{Rusu16progressive}, EWC \citep{Kirkpatrick17}, SI \citep{Zenke17}, LWF \citep{Li17}, ICARL \citep{Rebuffi16}, GEM \citep{Lopez-Paz17}, FearNet \citep{kemker18fearnet}, GDM \citep{Parisi18}, ExStream \citep{Hayes18MemoryEfficient}, Pure Rehearsal, GR \citep{Shin17}, MeRGAN \citep{wu2018memory} and AR1 \citep{Maltoni18}. Rehearsal and Generative Replay upper categories can be seen as a subset of replay strategies.}
  \label{fig:Venn}
\end{figure}

We show in Fig.~\ref{fig:Venn} some of the most popular and recent CL approaches divided into the above-introduced categories and their combinations.
In the diagram, we differentiate methods with \textit{rehearsal} (replay of explicitly stored training samples) from methods with \textit{generative replay} (replay of latent representations or the training samples).
Crucially, although an increasing number of methods have been proposed, there is no consensus on which training schemes and performance metrics are better to evaluate CL models.
Different sets of metrics have been proposed to evaluate CL performance on supervised and unsupervised learning tasks~(e.g.~\citep{Hayes18NewMetrics,Kemker17,Diaz18}).
In the absence of standardized metrics and evaluation schemes, it is unclear what it means to endow a method with CL capabilities.
In particular, a number of CL models still require large computational and memory resources that hinder their ability to learn in real time, or with a reasonable latency, from data streams.

It is also worth noting that, while a multitude of methods for each main category has been proposed, it is still difficult to find hybrid algorithmic solutions that can flexibly leverage the often orthogonal advantages of the three different approaches (i.e. architectural, regularization and replay), depending on the specific application needs and target efficiency-effectiveness trade-off. However, some evidence shows that effective biological continual learning systems (such as the human brain) make use of all these distinct functionalities. 

In this paper, we argue that in the near and long term future of lifelong learning machines \emph{we will witness a significantly growing interest in the development of hybrid continual learning algorithms} \citep{lomonaco2022cvpr} and we propose \texttt{ARR} as one of the first methodologies that practically implement such a vision. 

\section{ARR: Architect, Regularize and Replay}
\label{sec:arr}

\begin{figure}[t]
\centering
\includegraphics[width=0.7\columnwidth]{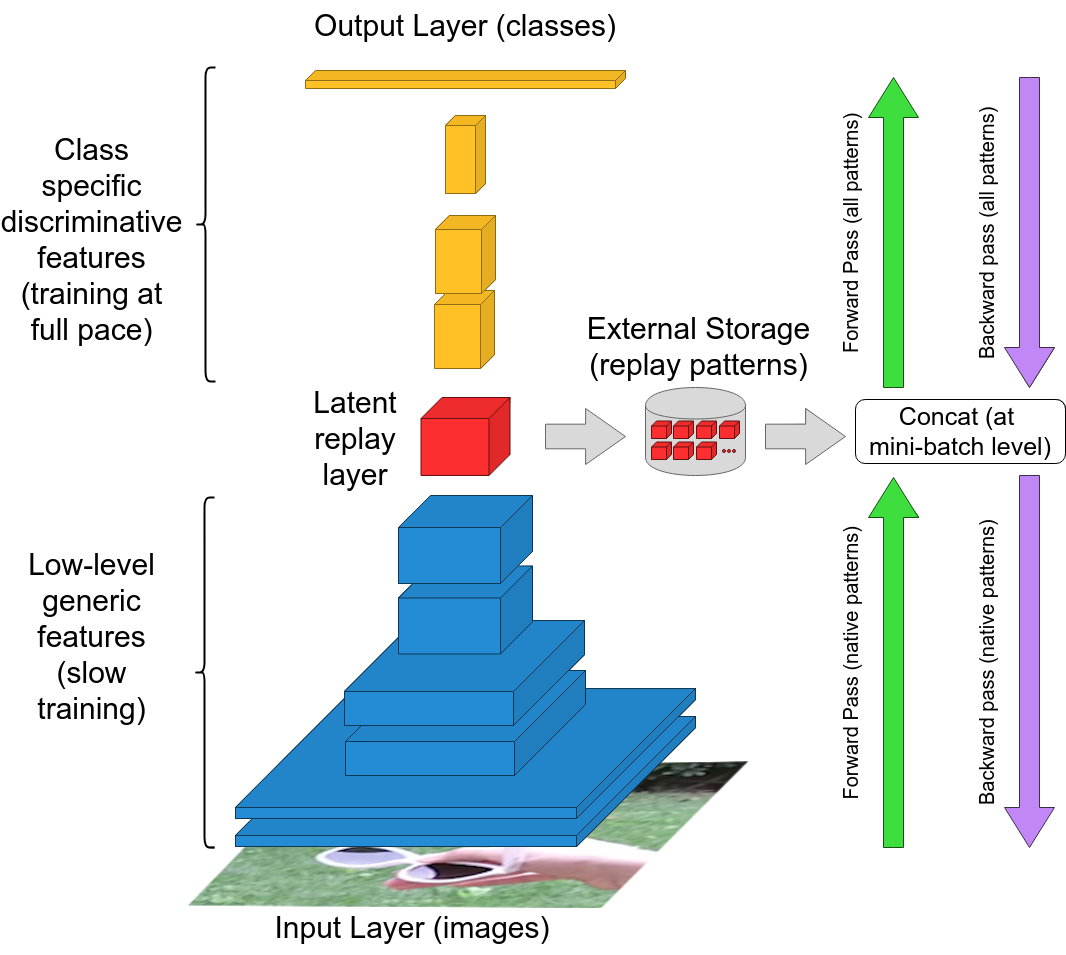}
\caption{Architectural diagram of ARR~\citep{pellegrini2019latent}.}
\label{fig:latent_replay}
\end{figure}

The \emph{Architect}, \emph{Regularize} and \emph{Replay} algorithm, \texttt{ARR} for short, is a flexible \emph{generalization} of the \texttt{AR1} algorithm and its variants (\texttt{CWR+}, \texttt{CWR*}, \texttt{AR1*}, \texttt{AR1Free}) \citep{lomonaco2020rehearsal, pellegrini2019latent}. \texttt{ARR}, with a proper initialization of its hyper-parameters, can be instantiated in the aforementioned algorithms based on the desired efficiency-efficacy trade-off \citep{ravaglia2020memory}. It can use pre-trained parameters as suggested by a consolidated trend in the field \citep{cossu2022continual} or start from a random initialization. The pseudo-code \ref{algo:arr} describes \texttt{ARR} in detail which is based on three main components: \emph{architectural}, \emph{regularization} and \emph{replay}.

\subsection{Architectural Component}

The core concept behind an architectural approach is to \emph{isolate} and preserve some parameters while adding new parameters in order to house new knowledge. \texttt{CWR+}, an evolution of \texttt{CWR} \citep{Lomonaco17} whose pseudo-code is reported in Algorithm 2 of \citep{maltoni2019} maintains two sets of weights for the output classification layer: $cw$ are the consolidated weights (for stability) used for inference and $tw$ the temporary weights (for plasticity) used for training; $cw$ are initialized to 0 before the first experience and then iteratively updated, while $tw$ are reset to 0 before each training experience.

In \citep{maltoni2019}, the authors proposed an extension of \texttt{CWR+} called \texttt{CWR*} which works both under \emph{Class-Incremental} \citep{Rebuffi17} and \emph{Class-Incremental with Repetition} settings \citep{cossu2022class}; in particular, under \emph{Class-Incremental with Repetition} the coming experiences include examples of both new and already encountered classes. For already known classes, instead of resetting weights to 0, consolidated weights are reloaded. Furthermore, in the consolidation step, a weighted sum is now used: the first term represents the weight of the past and the second term is the contribution from the current training experience. The weight $wpast_j$ used for the first term is proportional to the ratio $\frac{past_j}{cur_j}$, where $past_j$ is the total number of examples of class $j$ encountered in past experiences whereas $cur_j$ is their count in the current experience. In case of a large number of small non-i.i.d. training experiences, the weight for the most recent experiences may be too low thus hindering the learning process. In order to avoid this, a square root is used in order to smooth the final value of $wpast_j$.

\subsection{Regularization Component}

The well-known \emph{Elastic Weight Consolidation} (EWC) pure regularization approach \citep{Kirkpatrick17} controls forgetting by proportionally constraining the model weights based on their estimated importance with respect to previously encountered data distributions and tasks. To this purpose, in a classification approach, a regularization term is added to the conventional cross-entropy loss, where each weight $\theta_k$ of the model is pulled back to their optimal value $\theta_k^*$ with a strength $F_k$ proportional to their estimated importance for modeling past knowledge: 

\begin{equation}
L=L_{cross}(\cdot) + \frac{\lambda}{2} \cdot \sum_{k} F_k \cdot (\theta_k - \theta_k^*)^2.
\label{eq:loss}
\end{equation}

\emph{Synaptic Intelligence} (SI) \citep{Zenke17} is an equally known lightweight variant of EWC where, instead of updating the Fisher information $F$ at the end of each experience\footnote{In this paper, for the EWC and ARR implementations we use a single Fisher matrix updated over time, following the approach described in \citep{maltoni2019}.}, $F_k$ are obtained by integrating the loss over the weight trajectories exploiting information already available during gradient descent. For both approaches, the weight update rule corresponding to equation \ref{eq:loss} is:

\begin{equation}
\theta^{'}_k = \theta_k - \eta \cdot \frac{\partial L_{cross}(\cdot)}{\partial\theta_k} - \eta \cdot F_k \cdot (\theta_k - \theta_k^*)
\label{eq:update}
\end{equation}

\noindent where $\eta$ is the learning rate. This equation has two drawbacks. Firstly, the value of $\lambda$ must be carefully calibrated: in fact, if its value is too high the optimal value of some parameters could be overshoot, leading to divergence (see discussion in Section 2 of \citep{maltoni2019}). Secondly, two copies of all model weights must be maintained to store both $\theta_k$ and $\theta_k^*$, leading to double memory consumption for each weight.
To overcome the above problems, the authors of \citep{lomonaco2020rehearsal} propose to replace the update rule of equation \ref{eq:update} with:

\begin{equation}
 \theta^{'}_k = \theta_k - \eta \cdot (1 - \frac{F_k}{max_F}) \cdot \frac{\partial L_{cross}(\cdot)}{\partial\theta_k}
\end{equation}

\noindent where $max_F$ is the maximum value for weight importance (we clip to $max_F$ the $F_k$ values larger than $max_F$). Basically, the learning rate is reduced to 0 (i.e., complete freezing) for weights of highest importance ($F_k=max_F$) and maintained to $\eta$ for weights whose $F_k=0$.  It is worth noting that these two update rules work differently: the former still moves weights with high $F_k$ in the direction opposite to the gradient and then makes a step in direction of the past (optimal) values; the latter tends to completely freeze weights with high $F_k$. However, in the experiments conducted in \citep{lomonaco2020rehearsal}, the two approaches lead to similar results, and therefore the second one is preferable since it solves the aforementioned drawbacks.
Regularization of learning parameters can be enforced both on the low-level generic features as well as on the class-specific discriminative features as implemented in \texttt{AR1*}. However, for the sake of simplicity in \texttt{ARR} we consider only the application of such regularization terms to the last group, since freezing or slowly finetuning the low-level generic features already proved to be an effective strategy.

\begin{algorithm}[h]
\captionsetup{font=small}
\caption{ARR pseudocode: $\bar{\Theta}$ are the class-shared parameters of the representation layers; the notation  $cw[j]$ / $tw[j]$ is used to denote the groups of consolidated / temporary weights corresponding to class $j$. Note that this version continues to work under New Classes (NC), which is seen here as a special case of New Classes and Instances (NIC) \citep{Lomonaco17}; in fact, since in NC the classes in the current experience were never encountered before, the step at line 7 loads 0 values for classes in $\Dorig^i$ because $cw_j$ were initialized to 0 and in the consolidation step (line 13) $wpast_j$ values are always 0. The external random memory $RM$ is populated across the training experiences. Note that the amount $h$ of examples to add progressively decreases to maintain a nearly balanced contribution from the different training experiences, but no constraints are enforced to achieve a class-balancing. $\lambda$ is the regularization strength, $\alpha$ is the replay layer. The three input parameters default to 0 if omitted.}
\label{algo:arr}
\begin{algorithmic}[1]
\footnotesize
\Procedure{ARR}{$RM_{size}$, $\lambda$, $\alpha$}
\State $RM = \varnothing$, $cw[j]=0$ and $past_j=0 \;\ \forall j$
\State $\text{init } \bar{\Theta} \text{ randomly or from pre-trained model (e.g. on ImageNet)}$
\State \textbf{for each} $\text{training experience } e_i$:
\State \ \ \ \ $tw[j]=
    \begin{cases}
      cw[j], & \text{if class } j \text{ in } \Dorig^i\\
      0, & \text{otherwise}
    \end{cases}$
\State \ \ \ \ $mb_e = 
    \begin{cases}
      \frac{|\Dorig^i|}{(|\Dorig^i| + RM_{size}) / mb_{size}}, & \text{if } e_i > e_1 \\
      mb_{size}, & \text{otherwise}
    \end{cases}$
\State \ \ \ \ $mb_r = mb_{size} - mb_e$
\State \ \ \ \ \textbf{for each} epoch: 
\State \ \ \ \ \ \ \ Sample $mb_e$ examples from $\Dorig^i$ and $mb_r$ examples from $RM$
\State \ \ \ \ \ \ \ train the model on sampled data (replay data to be injected at leyer $\alpha$):
\State \ \ \ \ \ \ \ \ \ \ \ \textbf{if} $e_i=e_1$ learn both $\bar{\Theta}$ and $tw$
\State \ \ \ \ \ \ \ \ \ \ \ \textbf{else} learn $tw$ and $\bar{\Theta}$ with $\lambda$ to control forgetting.
\State \ \ \ \ \textbf{for each} class $j$ in $\Dorig^i$:
\State \ \ \ \ \ \ \ \ $wpast_j = \sqrt{\frac{past_j}{cur_j}}$, \text{where} $cur_j$ is the number of examples of class $j$ in $\Dorig^i$
\State \ \ \ \ \ \ \ \ $cw[j]=\frac{cw[j] \cdot wpast_j + (tw[j]-avg(tw))}{wpast_j + 1}$
\State \ \ \ \ \ \ \ \ $past_j = past_j + cur_j$
\State \ \ \ \ test the model by using $\bar{\Theta}$ and $cw$
\State \ \ \ \ $h = \dfrac{RM_{size}}{i}$
\State \ \ \ \ $R_{add} =$ random sampling $h$ examples from $\Dorig^i$
\State \ \ \ \ $R_{replace} = 
    \begin{cases}
      \varnothing & \text{if $i==1$}\\
      \text{random sample } h \text{ examples from } RM & \text{otherwise}
    \end{cases}$
\State \ \ \ \ $RM = (RM - R_{replace}) \cup R_{add}$
\EndProcedure
\end{algorithmic}
\end{algorithm}

\subsection{Replay Component}

In \citep{pellegrini2019latent, merlin2022practical} it was shown that a very simple rehearsal implementation (hereafter denoted as \emph{native rehearsal}), where for every training experience a random subset of the experience examples is added to the external storage to replace a (equally random) subset of the external memory, is not less effective than more sophisticated approaches such as iCaRL.  Therefore, in \citep{pellegrini2019latent} the authors opted for simplicity and compared the learning trend of \texttt{CWR*} and \texttt{AR1*} of a MobileNetV1\footnote{The network was pre-trained on ImageNet-1k.} trained with and without rehearsal on CORe50 NICv2 – 391 \citep{lomonaco2020rehearsal}. They used the same protocol and hyper-parameters introduced in \citep{Lomonaco2019} and a rehearsal memory of 1,500 examples. It is well evident from their study that even a moderate external memory (about 1.27\% of the total training set) is very effective to improve the accuracy of both approaches and to reduce the gap with the cumulative upper bound that, for this model, is $\sim$85\%.


In deep neural networks the layers close to the input (often denoted as representation layers) usually perform low-level feature extraction and, after a proper pre-training on a large dataset (e.g., ImageNet), their weights are quite stable and reusable across applications. On the other hand, higher layers tend to extract class-specific discriminant features and their tuning is often important to maximize accuracy.

A \emph{latent replay} (see Figure \ref{fig:latent_replay}) approach \citep{pellegrini2019latent} can then be formulated: instead of maintaining copies of input examples in the external memory in the form of raw data, we can store the \emph{activations volumes} at a given layer (denoted as \emph{latent replay layer}). To keep the representation stable and the stored activations valid we propose to slow down the learning at all the layers below the latent replay one and to leave the layers above free to learn at full pace. In the limit case where lower layers are completely frozen (i.e., slow-down to 0) latent replay is functionally equivalent to rehearsal from the input, but achieves a computational and storage saving thanks to the smaller fraction of examples that need to flow forward and backward across the entire network and the typical information compression that networks perform at higher layers.

In the general case where the representation layers are not completely frozen, the activations stored in the external memory may suffer from an \emph{aging effect} (i.e., as time passes they tend to increasingly deviate from the activations that the same pattern would produce if feed-forwarded from the input layer). However, if the training of these layers is sufficiently slow, the aging effect is not disruptive since the external memory has enough time to be updated with newly acquired examples. When latent replay is implemented with mini-batch SGD training: \emph{(i)} in the forward step, a concatenation is performed at the replay layer (on the mini-batch dimension) to join examples coming from the input layer with activations coming from the external storage; \emph{(ii)} the backward step is stopped just before the replay layer for the replay examples.

\section{Empirical Evaluation}
\label{sec:eval}

In order to empirically evaluate the overall quality and flexibility of \texttt{ARR}, we evaluate its performance on three commonly used continual learning benchmarks for computer vision classification tasks: \texttt{CIFAR-100} (Section \ref{subsec:cifar}), \texttt{CORe50} (Section \ref{subsec:core50}) and \texttt{ImageNet-1000} (Section \ref{subsec:imagenet}). Then, we provide a more in-depth analysis of the impact of the latent replay layer selection (Sec: \ref{subsec:replay_layer_selection}) and the memory size in terms of memorized activations volumes (Sec: \ref{subsec:replay_mem_size_selection}).

\subsection{CIFAR-100}
\label{subsec:cifar}

CIFAR-100 \citep{Krizhevsky09} is a well-known and largely used dataset for small ($32\times32$) natural image classification. It includes 100 classes containing 600 images each (500 training + 100 test). The default classification benchmark can be translated into a \emph{Class-Incremental} scenario (denoted as iCIFAR-100 by \citep{Rebuffi16}) by splitting the 100 classes into groups. In this paper, we consider groups of 10 classes thus obtaining 10 incremental experiences. 

\begin{figure}[!htb]
  \centering
  \includegraphics[width=\textwidth]{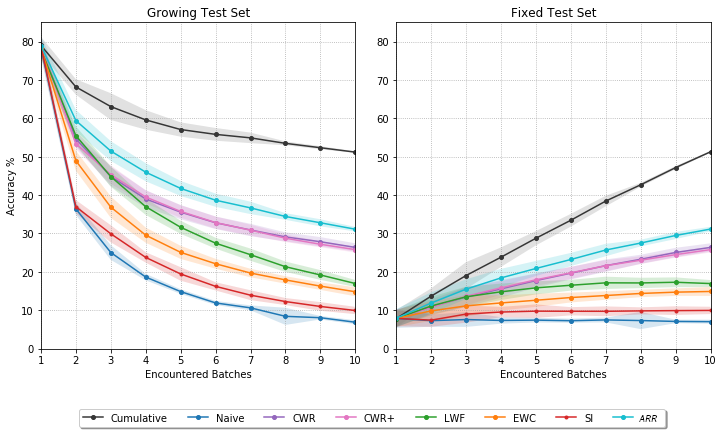}
  \caption{Accuracy on iCIFAR-100 with 10 experiences (10 classes per experience). Results are averaged on 10 runs: for all the strategies hyperparameters have been tuned on run 1 and kept fixed in the other runs. The experiment on the right, consistently with the CORe50 test protocol, considers a fixed test set including all the 100 classes, while on the left we include in the test set only the classes encountered so far (analogously to results reported in \citep{Rebuffi16}). Colored areas represent the standard deviation of each curve. Better viewed in color. \citep{maltoni2019}}
  \label{fig:icifar_sit_results}
\end{figure}

The CNN model used for this experiment is the same used by \citep{Zenke17} for experiments on CIFAR-10/100 Split \citep{maltoni2019}. It consists of 4 convolutional + 2 fully connected layers; details are available in Appendix A of \citep{Zenke17}. The model was pre-trained on CIFAR-10 \citep{Krizhevsky09}. Figure \ref{fig:icifar_sit_results} compares the accuracy of the different approaches on iCIFAR-100. The results suggest that:

\begin{itemize}
	\item Unlike the \texttt{Naïve} approach, \emph{Learning without Forgetting} (\texttt{LWF}) \citep{Li17} and \emph{Elastic Weights Consolidation} (\texttt{EWC}) provide some robustness against forgetting, even if in this incremental scenario their performance is not satisfactory. \texttt{SI}, when used in isolation, is quite unstable and performs worse than \texttt{LWF} and \texttt{EWC}.
	\item The accuracy improvement of \texttt{CWR+} over \texttt{CWR} is here very small because the experiences are balanced (so weight normalization is not required) and the CNN initialization for the last level weights was already very close to 0 (we used the authors’ default setting of a Gaussian with std = 0.005).
	\item \texttt{ARR {\small($\lambda=4.0e5$)}} consistently outperforms all the other approaches.
\end{itemize}


It is worth noting that both the experiments reported in Figure \ref{fig:icifar_sit_results} (i.e., an expanding (left) and fixed (right) test set, from left to right) lead to the same conclusions in terms of relative ranking among approaches. However, we believe that a fixed test set allows to better appreciate the incremental learning trend and its peculiarities (saturation, forgetting, etc.) because the classification complexity (which is proportional to the number of classes) remains constant across the experiences. For example, in the right graph it can be noted that \texttt{SI}, \texttt{EWC} and \texttt{LWF} learning capacities tend to saturate after 6-7 experiences while \texttt{CWR}, \texttt{CWR+} and \texttt{ARR} continue to grow; the same information is not evident on the left because of the underlying negative trend due to the increasing problem complexity.

Finally note that absolute accuracy on \texttt{iCIFAR-100} cannot be directly compared with \citep{Rebuffi16} because the CNN model used in \citep{Rebuffi16} is a ResNet-32, which is much more accurate than the model here used: on the full training set the model here used achieves about 51\% accuracy while ResNet-32 about 68.1\%.


\subsection{CORe50}
\label{subsec:core50}

\begin{figure}[t]
\centering
\includegraphics[width=0.5\columnwidth]{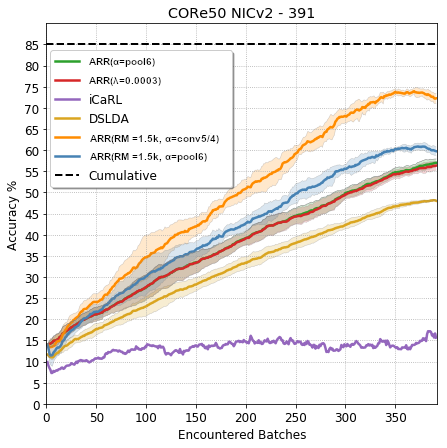}
\caption{Accuracy results on the CORe50 NICv2 – 391 benchmark of \texttt{ARR{\small($\alpha=pool6$)}}, \texttt{ARR{\small($\lambda=0.0003$)}}, DSLDA, iCaRL, \texttt{ARR{\small($RM_{size}=1500,\alpha=conv5\_4$)}}, \texttt{ARR{\small($RM_{size}=1500,\alpha=pool6$)}}. Results are averaged across 10 runs in which the order of the experiences is randomly shuffled. Colored areas indicate the standard deviation of each curve. As an exception, iCaRL was trained only on a single run given its extensive run time ($\sim$14 days).}
\label{fig:comparison}
\end{figure}

While the accuracy improvement of the proposed approach w.r.t. the state-of-the-art rehearsal-free techniques have been already discussed in the previous section, further comparison with other state-of-the-art continual learning techniques on CORe50 may be beneficial for better appreciating its practical impact and advantages in real-world continual learning scenarios and longer sequences of experiences.
In particular, while \texttt{ARR} and \texttt{ARR{\small($\alpha=pool6$)}} have been already proven to be substantially better than \texttt{LWF} and \texttt{EWC} on the NICv2 - 391 benchmark \citep{lomonaco2020rehearsal}, a comparison with \texttt{iCaRL}\citep{Rebuffi17}, one of the best know rehearsal-based techniques, is worth to be considered. 

Unfortunately, \texttt{iCaRL} was conceived for \emph{Class-Incremental} scenarios and its porting to \emph{Class-Incremental with Repetition} (whose experiences also include examples of know classes) is not trivial. To avoid subjective modifications, the authors of \citep{lomonaco2020rehearsal} started from the code shared by its original authors and emulated a \emph{Class-Incremental with Repetition} setting by: \emph{(i)} always creating new virtual classes from examples in the coming experiences; \emph{(ii)} fusing virtual classes together when evaluating accuracies. For example, let us suppose to encounter 300 examples of class 5 in experience 2 and other 300 examples of the same class in experience 7; while two virtual classes are created by \texttt{iCaRL} during training, when evaluating accuracy both classes point to the real class 5.
Such \texttt{iCaRL} implementation, with an external memory of 8000 examples (much more than the 1500 used by the proposed latent replay, but in line with the settings proposed in the original paper \citep{Rebuffi16}), was run on NICv2 - 391, but we were not able to obtain satisfactory results. In Figure \ref{fig:comparison} we report the \texttt{iCaRL} accuracy over time and compare it with \texttt{ARR{\small($RM_{size}=1500, \alpha=conv5\_4/dw$)}}, \texttt{ARR{\small($RM_{size}=1500, \alpha=pool6$)}} as well as the top three performing rehearsal-free strategies introduced before: \texttt{ARR{\small($\alpha=pool6$)}}, \texttt{ARR{\small($\lambda=0.0003$)}} and \texttt{DSLDA}. While \texttt{iCaRL} exhibits better performance than \texttt{LWF} and \texttt{EWC} (as reported in \citep{Lomonaco2019}), it is far from \texttt{DSLDA}, \texttt{ARR{\small($\alpha=pool6$)}} and \texttt{ARR{\small($\lambda=0.0003$)}}.

Furthermore, when the algorithm has to deal with a so large number of classes (including virtual ones) and training experiences its efficiency becomes very low (as also reported in \citep{maltoni2019}). In Table 1 of \citep{lomonaco2020rehearsal} the total run time (training and testing), memory overhead and accuracy difference with respect to the cumulative upper bound are reported. We believe \texttt{ARR{\small($RM_{size}=1500, \alpha=conv5\_4/dw$)}} represents a good trade-off in terms of efficiency-efficacy with a limited computational-memory overhead and only a $\sim$13\% accuracy gap from the cumulative upper bound. For \texttt{iCaRL} the total training time was $\sim$14 days compared to a training time of less than $\sim$1 hour for the other learning algorithms on a single GPU.

\subsection{ImageNet-1000}
\label{subsec:imagenet}

In order to further validate the \texttt{ARR} algorithm scalability the authors of \citep{graffieti2022generative} performed a test on a competitive benchmark such as ImageNet-1000, following the \emph{Class-Incremental} benchmark proposed by \citep{masana2020class}, which is composed of 25 experiences, each of them containing 40 classes. The benchmark is particularly challenging due to the large number of classes (1,000), the incremental nature of the task (with 25 experiences), and the data dimensionality of $224 \times 224$ (as with ImageNet protocol). 

In this experiment, \citep{graffieti2022generative} tested \texttt{ARR} against both regularization-based methods \citep{dhar2019learning, Kirkpatrick17, Li17} and replay-based approaches \citep{belouadah2019il2m, castro2018end, chaudhry2018riemannian, hou2019learning, Rebuffi17, wu2019large}. They used the same classifier (ResNet-18) and the same memory size for all the tested methods (20,000 examples, 20 per class); for the regularization-based approaches, the replay is added as an additional mechanism. 

For \texttt{ARR}, they trained the model with an SGD optimizer. For the first experience, the algorithm was tuned with an aggressive learning rate of $0.1$ with momentum of $0.9$ and weight decay of $10^{-4}$. Then, the initial learning rate was multiplied by $0.1$ every $15$ epochs. The model was trained for a total of $45$ epochs, using a batch size of $128$. For all the subsequent experiences SGD with a learning rate of $5 \cdot 10^{-3}$ for the feature extractor's parameters $\phi$ and $5 \cdot 10^{-2}$ for the classifier's parameters $\psi$ were used. The model was trained for $32$ epochs for each experience, employing a learning rate scheduler that decreases the learning rate as the number of experiences progresses. This was done to protect old knowledge against new knowledge when the former is more abundant than the latter. As in the first experience, the batch size was set to $128$, composed of $92$ examples from the current experience and $36$ randomly sampled (without replacement) from the replay memory.

The results are shown in~\autoref{tab:imagenet_40_25}. Replay-based methods exhibit the best performance, with \texttt{iCaRL} and \texttt{BiC} exceeding a final accuracy of 30\%. \texttt{ARR{\small($RM_{size}=1500, \alpha=pool6$)}} outperforms all the baselines (33.1\%) achieving state-of-the-art performance on this challenging benchmark, and proving the advantage of flexible hybrid continual learning approaches. However, considering that top-1 ImageNet accuracy for a ResNet-18 when trained on the entire dataset is 69.76\%\footnote{Accuracy taken from the torchvision official page: \url{https://pytorch.org/vision/stable/models.html}}, even for the best methods the accuracy gap in the continual learning setup is very large. This suggests that continual learning, especially in complex scenarios with a large number of classes and high dimensional data, is far to be solved, and further research should be devoted to this field. 

\begin{table}[]
    \centering
    \begin{tabular}{c c }
    \toprule
        \textbf{Method} & \textbf{Final Accuracy} \\
        \midrule
        Fine Tuning (Naive) & 27.4 \\
        EWC-E \citep{Kirkpatrick17} & 28.4 \\
        RWalk \citep{Chaudhry18} & 24.9 \\ 
        LwM \citep{Dhar19} & 17.7 \\
        LwF \citep{Li19} & 19.8 \\
        iCaRL \citep{Rebuffi17} & 30.2 \\
        EEIL \citep{castro2018end} &  25.1 \\ 
        LUCIR \citep{hou2019learning} & 20.1 \\
         IL2M \citep{belouadah2019il2m} & 29.7 \\ 
         BiC \citep{wu2019large} & 32.4 \\
         \textbf{ARR} \citep{maltoni2019} & \textbf{33.1} \\
    \bottomrule
    \vspace{1px}
    \end{tabular}
    \caption{Final accuracy on ImageNet-1000 following the benchmark of \citep{masana2020class} with 25 experiences composed of 40 classes each. For each method, a replay memory of 20,000 examples is used (20 per class at the end of training). Results for other methods reported from~\citep{masana2020class}.}
    \label{tab:imagenet_40_25}
\end{table}

\subsection{Replay Layer Selection}
\label{subsec:replay_layer_selection}

\begin{figure}[h]
\centering
\includegraphics[width=0.5\columnwidth]{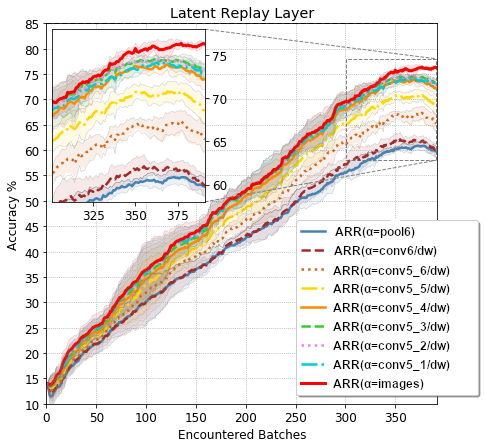}
\caption{\texttt{ARR} with latent replay ($RM_{size}=1500$) for different choices of the latent replay layer. Setting the replay layer at the ``images'' layer corresponds to native rehearsal. The saturation effect which characterizes the last training experiences is due to the data distribution in NICv2 – 391 (see \citep{Lomonaco2019}): in particular, the lack of new instances for some classes (that already introduced all their data) slows down the accuracy trend and intensifies the effect of activations aging.}
\label{fig:latent_layer_diff}
\end{figure}

In Figure \ref{fig:latent_layer_diff} we report the accuracy of \texttt{ARR{\small($RM_{size}=1500, \alpha=\dots$)}} for different choices of the rehearsal layer $\alpha$ for the CORe50 experiment. As expected, when the replay layer is pushed down the corresponding accuracy increases, proving that a continual tuning of the representation layers is important. However, after \texttt{conv5\_4/dw} there is a sort of saturation and the model accuracy is no longer improving. The residual gap ($\sim$4\%) with respect to native rehearsal is not due to the weights freezing of the lower part of the network but to the aging effect introduced above. This can be simply proved by implementing an ``intermediate'' approach that always feeds the replay pattern from the input and stops the backward at \texttt{conv5\_4}: such an intermediate approach achieved an accuracy at the end of the training very close to the native rehearsal (from raw data). We believe that the accuracy drop due to the aging effect can be further reduced with better tuning of \emph{Batch Re-Normalization} (BRN) hyper-parameters and/or with the introduction of a scheduling policy making the global moment mobile windows wider as the continual learning progresses (i.e., more plasticity in the early stages and more stability later); however, such fine optimization is application specific and beyond the scope of this study.

To better evaluate the latent replay with respect to the native rehearsal we report in Table \ref{tab:tradeoff} the relevant dimensions: \emph{(i)} computation refers to the percentage cost in terms of ops of a partial forward (from the latent replay layer on) relative to a full forward step from the input layer; \emph{(ii)} pattern size is the dimensionality of the pattern to be stored in the external memory (considering that we are using a MobileNetV1 with 128$\times$128$\times$3 inputs to match CORe50 image size); \emph{(iii)} accuracy and $\Delta$ accuracy quantify the absolute accuracy at the end of the training and the gap with respect to a native rehearsal, respectively. For example, \texttt{conv5\_4/dw} exhibits an interesting trade-off because the computation is about 32\% of the native rehearsal one, the storage is reduced to 66\% (more on this point in subsection \ref{subsec:replay_mem_size_selection}) and the accuracy drop is mild (5.07\%). \texttt{ARR{\small($RM_{size}=1500, \alpha=pool6$)}} has a really negligible computational cost (0.027\%) with respect to native rehearsal and still provides an accuracy improvement of $\sim$4\% w.r.t. the non-rehearsal case ($\sim$60\% vs $\sim$56\% as it is possible to see from Figure \ref{fig:latent_layer_diff} and Figure \ref{fig:mem_size}, respectively).

\begin{table*}[th]
  \caption{Computation, storage, and accuracy trade-off with Latent Replay at different layers of a MobileNetV1 ConvNet trained continually on NICv2 – 391 with $RM_{size} = 1500$.}
  \label{tab:tradeoff}
  \centering
  \small
  \renewcommand{\arraystretch}{1.05}
  \setlength\tabcolsep{0.1cm}
  \begin{tabular}{p{1.5cm}cccc}
    \toprule
    \thead[l]{Layer} & \thead{Computation \% \\ vs Native Rehearsal} & \thead{Example Size} & \thead{Final Accuracy} \% & \thead{$\Delta$ Accuracy \% \\ vs Native Rehearsal}\\
    \midrule
    Images & 100.00\% & 49152 & 77.30\% & 0.00\% \\
    conv5\_1/dw & 59.261\% & 32768 & 72.82\% &-4.49\%\\
    conv5\_2/dw & 50.101\% & 32768 & 73.21\% &-4.10\%\\
    conv5\_3/dw & 40.941\% & 32768 & 73.22\% & -4.09\% \\
    \textbf{conv5\_4/dw} & \textbf{31.781\%} & \textbf{32768} & \textbf{72.24\%} & \textbf{-5.07\%} \\
    conv5\_5/dw & 22.621\% & 32768 & 68.59\% & -8.71\% \\
    conv5\_6/dw & 13.592\% & 8192 & 65.24\% & -12.06\% \\
    conv6/dw & 9.012\% & 16384 & 59.89\% & -17.42\% \\
    pool6 & 0.027\% & 1024 & 59.76\% & -17.55\% \\
    \bottomrule
  \end{tabular}
\end{table*}

\subsection{Replay Memory Size Selection}
\label{subsec:replay_mem_size_selection}

\begin{figure*}[ht]
\centering
\includegraphics[width=\textwidth]{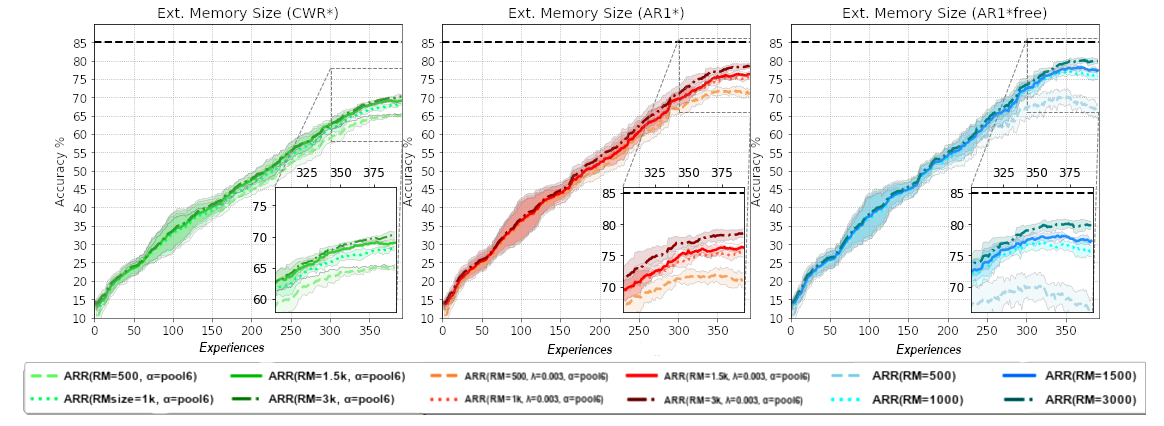}
\caption{Comparison of main \texttt{ARR} configurations on CORe50 NICv2 – 391 with different external memory sizes ($RM_{size} = 500, 1000, 1500$ and $3000$ examples).}
\label{fig:mem_size}
\end{figure*}

To understand the influence of the external memory size we repeated the experiment with different $RM_{size}$ values: 500, 1,000, 1,500, 3,000. The results are shown in Figure \ref{fig:mem_size}: it is worth noting that increasing the rehearsal memory leads to better accuracy for all the algorithms, but the gap between 1500 and 3000 is not large and we believe 1500 is a good trade-off for this dataset. \texttt{ARR{\small($RM_{size}=\dots$)}} works slightly better than \texttt{ARR{\small($RM_{size}=\dots, \lambda=0.003$)}} when a sufficient number of rehearsal examples are provided but, as expected, accuracy is worse with light (i.e. $RM_{size} = 500$) or no rehearsal.

It is worth noting that the best \texttt{ARR} configuration in Figure \ref{fig:mem_size}, i.e. \texttt{ARR{\small($RM_{size}=3000$)}}, is only 5\% worse than the cumulative upper bound and a better parametrization and exploitation of the rehearsal memory could further reduce this gap.

\section{ARR Implementation in Avalanche}
\label{sec:avalanche}

The \emph{Architect, Regularize} and \emph{Replay} (ARR) method we proposed in this paper is the result of a comprehensive re-formalization of different variants and improvements proposed over the last few years starting from \citep{Lomonaco17,maltoni2019}. Original implementations of such methods (\texttt{CWR}, \texttt{CWR+}, \texttt{CWR*}, \texttt{AR1}, \texttt{AR1*} and \texttt{AR1*} with \emph{Latent Replay}) exist in Caffe and PyTorch. However, given their diversity, it is quite difficult to move from one implementation to the other and apply them to settings and scenarios even slightly different from the ones on which they have been proposed.

In order to exploit the general applicability and flexibility of the \texttt{ARR} method, we decided to re-implement it directly in Avalanche \citep{lomonaco2021avalanche}. \emph{Avalanche}, an open-source (MIT licensed) end-to-end library for continual learning based on PyTorch, was devised to provide a shared and collaborative codebase for fast prototyping, training, and evaluation of continual learning algorithms.

\begin{figure}[h]
\centering
\includegraphics[width=\columnwidth]{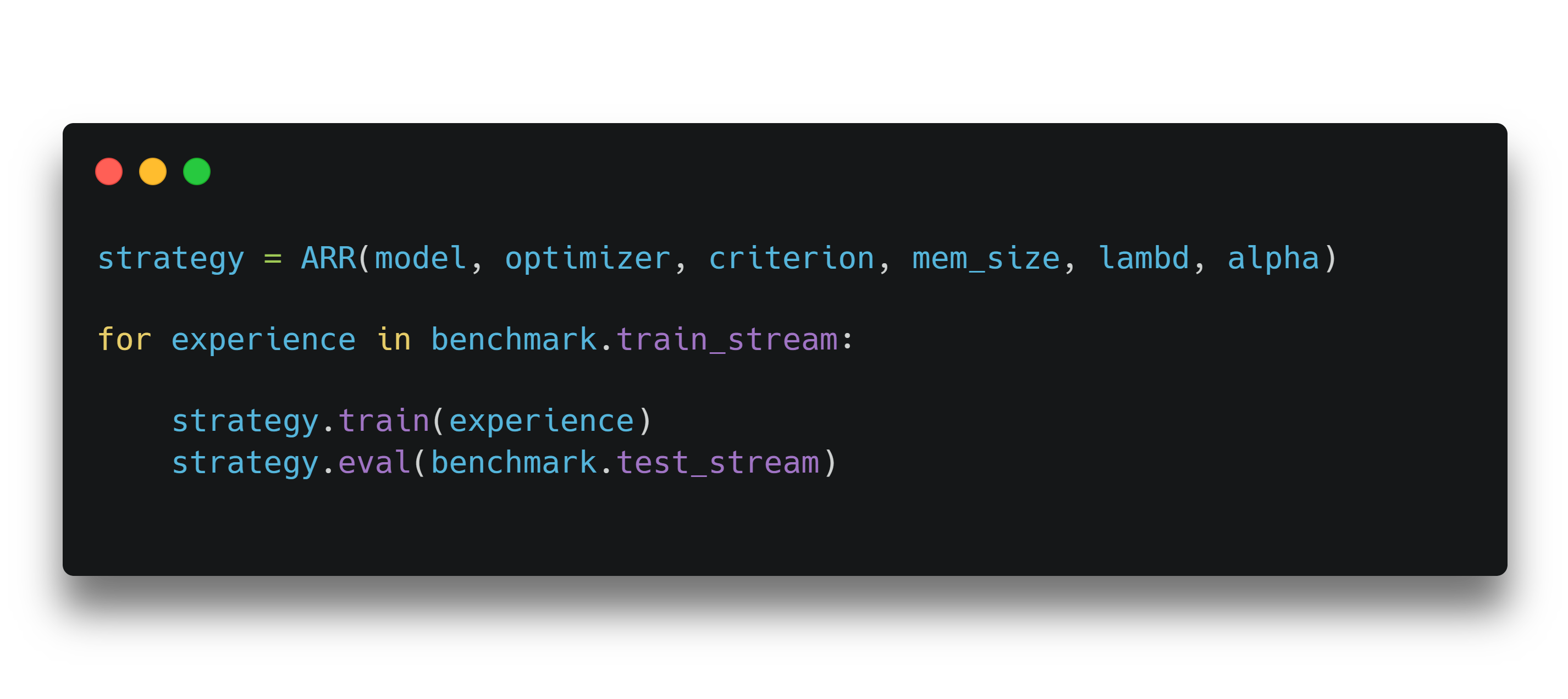}
\caption{\texttt{ARR} implementation in Avalanche. Given a set of hyper-parameters \texttt{ARR} can be instantiated and properly configured to be tested on a large set of benchmarks already available in Avalanche.}
\label{fig:arr-avalanche}
\end{figure}

Thanks to the Avalanche portable implementation (soon to be integrated into the next stable version of the library), \texttt{ARR} can be configured to reproduce the experiments presented in this paper (Fig. \ref{fig:arr-avalanche}, conform to the previously proposed strategies (e.g. \texttt{AR1*}, \texttt{CWR*}, etc.) as well as being ready to be tested on a large set of benchmarks already available in \texttt{Avalanche} or that can be easily added to the library.

\section{Conclusion}

In this paper we showed that \texttt{ARR} is a flexible, effective and efficient technique to continually learn new classes and new instances of known classes even from small and non i.i.d. experiences. \texttt{ARR}, instantiated with latent replay, is indeed able to learn efficiently and, at the same time, the achieved accuracy is not far from the cumulative upper bound (about 5\% in some cases).
The computation-storage-accuracy trade-off can be defined according to both the target application and the available resources so that even edge devices with no GPUs can learn continually. Moreover, \texttt{ARR} can be easily extended to support more sophisticated replay memory management strategies (also to contrast the \emph{aging effect}) and even be coupled with a generative model trained in the loop and capable of providing pseudo-activations volumes on demand as initially showed in \citep{graffieti2022generative}.

\section*{References}

\bibliographystyle{plain} 
\bibliography{biblio} 

\begin{thebibliography}{10}

\bibitem{belouadah2019il2m}
Eden Belouadah and Adrian Popescu.
\newblock Il2m: Class incremental learning with dual memory.
\newblock In {\em Proceedings of the IEEE/CVF International Conference on
  Computer Vision}, pages 583--592, 2019.

\bibitem{carta2021ex}
Antonio Carta, Andrea Cossu, Vincenzo Lomonaco, and Davide Bacciu.
\newblock Ex-model: Continual learning from a stream of trained models.
\newblock {\em arXiv preprint arXiv:2112.06511}, 2021.

\bibitem{castro2018end}
Francisco~M Castro, Manuel~J Mar{\'\i}n-Jim{\'e}nez, Nicol{\'a}s Guil, Cordelia
  Schmid, and Karteek Alahari.
\newblock End-to-end incremental learning.
\newblock In {\em Proceedings of the European conference on computer vision
  (ECCV)}, pages 233--248, 2018.

\bibitem{chaudhry2018riemannian}
Arslan Chaudhry, Puneet~K Dokania, Thalaiyasingam Ajanthan, and Philip~HS Torr.
\newblock Riemannian walk for incremental learning: Understanding forgetting
  and intransigence.
\newblock In {\em Proceedings of the European Conference on Computer Vision
  (ECCV)}, pages 532--547, 2018.

\bibitem{Chaudhry18}
Arslan Chaudhry, Puneet~K Dokania, Thalaiyasingam Ajanthan, and Philip~HS Torr.
\newblock Riemannian walk for incremental learning: Understanding forgetting
  and intransigence.
\newblock In {\em ECCV}, 2018.

\bibitem{cossu2022class}
Andrea Cossu, Gabriele Graffieti, Lorenzo Pellegrini, Davide Maltoni, Davide
  Bacciu, Antonio Carta, and Vincenzo Lomonaco.
\newblock Is class-incremental enough for continual learning?
\newblock {\em Frontiers in Artificial Intelligence}, 5, 2022.

\bibitem{cossu2022continual}
Andrea Cossu, Tinne Tuytelaars, Antonio Carta, Lucia Passaro, Vincenzo
  Lomonaco, and Davide Bacciu.
\newblock Continual pre-training mitigates forgetting in language and vision.
\newblock {\em arXiv preprint arXiv:2205.09357}, 2022.

\bibitem{dhar2019learning}
Prithviraj Dhar, Rajat~Vikram Singh, Kuan-Chuan Peng, Ziyan Wu, and Rama
  Chellappa.
\newblock Learning without memorizing.
\newblock In {\em Proceedings of the IEEE/CVF conference on computer vision and
  pattern recognition}, pages 5138--5146, 2019.

\bibitem{Dhar19}
Prithviraj Dhar, Rajat~Vikram Singh, Kuan-Chuan Peng, Ziyan Wu, and Rama
  Chellappa.
\newblock Learning without memorizing.
\newblock In {\em The IEEE Conference on Computer Vision and Pattern
  Recognition (CVPR)}, June 2019.

\bibitem{Diaz18}
Natalia D{\'i}az-Rodr{\'i}guez, Vincenzo Lomonaco, David Filliat, and Davide
  Maltoni.
\newblock {Don't forget, there is more than forgetting: new metrics for
  Continual Learning}.
\newblock In {\em {Workshop on Continual Learning, NeurIPS 2018 (Neural
  Information Processing Systems}}, Montreal, Canada, December 2018.

\bibitem{graffieti2022generative}
Gabriele Graffieti, Davide Maltoni, Lorenzo Pellegrini, and Vincenzo Lomonaco.
\newblock Generative negative replay for continual learning.
\newblock {\em arXiv preprint arXiv:2204.05842}, 2022.

\bibitem{Hayes18NewMetrics}
T.~L. Hayes, R.~Kemker, N.~D. Cahill, and C.~Kanan.
\newblock New metrics and experimental paradigms for continual learning.
\newblock In {\em 2018 IEEE/CVF Conference on Computer Vision and Pattern
  Recognition Workshops (CVPRW)}, pages 2112--21123, June 2018.

\bibitem{Hayes18MemoryEfficient}
Tyler~L. Hayes, Nathan~D. Cahill, and Christopher Kanan.
\newblock Memory efficient experience replay for streaming learning.
\newblock {\em 2019 International Conference on Robotics and Automation
  (ICRA)}, pages 9769--9776, 2018.

\bibitem{hou2019learning}
Saihui Hou, Xinyu Pan, Chen~Change Loy, Zilei Wang, and Dahua Lin.
\newblock Learning a unified classifier incrementally via rebalancing.
\newblock In {\em Proceedings of the IEEE/CVF Conference on Computer Vision and
  Pattern Recognition}, pages 831--839, 2019.

\bibitem{kemker18fearnet}
Ronald Kemker and Christopher Kanan.
\newblock Fearnet: Brain-inspired model for incremental learning.
\newblock In {\em International Conference on Learning Representations}, 2018.

\bibitem{Kemker17}
Ronald Kemker, Marc McClure, Angelina Abitino, Tyler~L. Hayes, and Christopher
  Kanan.
\newblock Measuring catastrophic forgetting in neural networks.
\newblock In {\em AAAI}, 2017.

\bibitem{Kirkpatrick17}
James Kirkpatrick, Razvan Pascanu, Neil Rabinowitz, Joel Veness, Guillaume
  Desjardins, Andrei~A Rusu, Kieran Milan, John Quan, Tiago Ramalho, Agnieszka
  Grabska-Barwinska, et~al.
\newblock Overcoming catastrophic forgetting in neural networks.
\newblock {\em Proc. of the national academy of sciences}, 2017.

\bibitem{Krizhevsky09}
Alex Krizhevsky, Geoffrey Hinton, et~al.
\newblock Learning multiple layers of features from tiny images.
\newblock Technical report, Citeseer, 2009.

\bibitem{lesort:hal-02381343}
Timoth{\'e}e Lesort, Vincenzo Lomonaco, Andrei Stoian, Davide Maltoni, David
  Filliat, and Natalia D{\'i}az-Rodr{\'i}guez.
\newblock {Continual Learning for Robotics: Definition, Framework, Learning
  Strategies, Opportunities and Challenges}.
\newblock {\em {Information Fusion}}, December 2019.

\bibitem{Li19}
Xilai Li, Yingbo Zhou, Tianfu Wu, Richard Socher, and Caiming Xiong.
\newblock Learn to grow: {A} continual structure learning framework for
  overcoming catastrophic forgetting.
\newblock {\em CoRR}, abs/1904.00310, 2019.

\bibitem{Li17}
Zhizhong Li and Derek Hoiem.
\newblock Learning without forgetting.
\newblock {\em IEEE Transactions on Pattern Analysis and Machine Intelligence},
  2017.

\bibitem{lomonaco2018thesis}
Vincenzo Lomonaco.
\newblock {\em {Continual Learning with Deep Architectures}}.
\newblock Phd thesis, University of Bologna, 2019.

\bibitem{lomonaco2019continual}
Vincenzo Lomonaco, Karan Desai, Eugenio Culurciello, and Davide Maltoni.
\newblock Continual reinforcement learning in 3d non-stationary environments.
\newblock {\em arXiv preprint arXiv:1905.10112}, 2019.

\bibitem{Lomonaco17}
Vincenzo Lomonaco and Davide Maltoni.
\newblock {CORe50: a New Dataset and Benchmark for Continuous Object
  Recognition}.
\newblock In Sergey Levine, Vincent Vanhoucke, and Ken Goldberg, editors, {\em
  Proceedings of the 1st Annual Conference on Robot Learning}, volume~78 of
  {\em Proceedings of Machine Learning Research}, pages 17--26. PMLR, 13--15
  Nov 2017.

\bibitem{Lomonaco2019}
Vincenzo Lomonaco, Davide Maltoni, and Lorenzo Pellegrini.
\newblock {Fine-Grained Continual Learning}.
\newblock {\em arXiv preprint arXiv: 1907.03799}, pages 1--14, 2019.

\bibitem{lomonaco2020rehearsal}
Vincenzo Lomonaco, Davide Maltoni, and Lorenzo Pellegrini.
\newblock Rehearsal-free continual learning over small non-iid batches.
\newblock In {\em CVPR Workshops}, volume~1, page~3, 2020.

\bibitem{lomonaco2021avalanche}
Vincenzo Lomonaco, Lorenzo Pellegrini, Andrea Cossu, Antonio Carta, Gabriele
  Graffieti, Tyler~L Hayes, Matthias De~Lange, Marc Masana, Jary Pomponi,
  Gido~M Van~de Ven, et~al.
\newblock Avalanche: an end-to-end library for continual learning.
\newblock In {\em Proceedings of the IEEE/CVF Conference on Computer Vision and
  Pattern Recognition}, pages 3600--3610, 2021.

\bibitem{lomonaco2022cvpr}
Vincenzo Lomonaco, Lorenzo Pellegrini, Pau Rodriguez, Massimo Caccia, Qi~She,
  Yu~Chen, Quentin Jodelet, Ruiping Wang, Zheda Mai, David Vazquez, et~al.
\newblock Cvpr 2020 continual learning in computer vision competition:
  Approaches, results, current challenges and future directions.
\newblock {\em Artificial Intelligence}, 303:103635, 2022.

\bibitem{Lopez-Paz17}
David Lopez-Paz and Marc-Aurelio Ranzato.
\newblock Gradient episodic memory for continual learning.
\newblock In I.~Guyon, U.~V. Luxburg, S.~Bengio, H.~Wallach, R.~Fergus,
  S.~Vishwanathan, and R.~Garnett, editors, {\em Advances in Neural Information
  Processing Systems 30}, pages 6467--6476. Curran Associates, Inc., 2017.

\bibitem{maltoni2019}
Davide Maltoni and Vincenzo Lomonaco.
\newblock {Continuous learning in single-incremental-task scenarios}.
\newblock {\em Neural Networks}, 116:56--73, aug 2019.

\bibitem{Maltoni18}
Davide Maltoni and Vincenzo Lomonaco.
\newblock Continuous learning in single-incremental-task scenarios.
\newblock {\em Neural Networks}, 116:56 -- 73, 2019.

\bibitem{masana2020class}
Marc Masana, Xialei Liu, Bartlomiej Twardowski, Mikel Menta, Andrew~D Bagdanov,
  and Joost van~de Weijer.
\newblock Class-incremental learning: survey and performance evaluation on
  image classification.
\newblock {\em arXiv preprint arXiv:2010.15277}, 2020.

\bibitem{merlin2022practical}
Gabriele Merlin, Vincenzo Lomonaco, Andrea Cossu, Antonio Carta, and Davide
  Bacciu.
\newblock Practical recommendations for replay-based continual learning
  methods.
\newblock {\em arXiv preprint arXiv:2203.10317}, 2022.

\bibitem{parisi2020online}
German~I Parisi and Vincenzo Lomonaco.
\newblock Online continual learning on sequences.
\newblock In {\em Recent Trends in Learning From Data}, pages 197--221.
  Springer, 2020.

\bibitem{Parisi18}
German~I. Parisi, Jun Tani, Cornelius Weber, and Stefan Wermter.
\newblock Lifelong learning of spatiotemporal representations with dual-memory
  recurrent self-organization.
\newblock {\em Frontiers in Neurorobotics}, 12:78, 2018.

\bibitem{pellegrini2019latent}
Lorenzo Pellegrini, Gabrile Graffieti, Vincenzo Lomonaco, and Davide Maltoni.
\newblock Latent replay for real-time continual learning.
\newblock {\em arXiv preprint arXiv:1912.01100}, 2019.

\bibitem{ravaglia2020memory}
Leonardo Ravaglia, Manuele Rusci, Alessandro Capotondi, Francesco Conti,
  Lorenzo Pellegrini, Vincenzo Lomonaco, Davide Maltoni, and Luca Benini.
\newblock Memory-latency-accuracy trade-offs for continual learning on a risc-v
  extreme-edge node.
\newblock In {\em 2020 IEEE Workshop on Signal Processing Systems (SiPS)},
  pages 1--6. IEEE, 2020.

\bibitem{Rebuffi16}
S.~{Rebuffi}, A.~{Kolesnikov}, G.~{Sperl}, and C.~H. {Lampert}.
\newblock icarl: Incremental classifier and representation learning.
\newblock In {\em 2017 IEEE Conference on Computer Vision and Pattern
  Recognition (CVPR)}, pages 5533--5542, July 2017.

\bibitem{Rebuffi17}
Sylvestre-alvise Rebuffi, Alexander Kolesnikov, Georg Sperl, and Christoph~H
  Lampert.
\newblock {iCaRL: Incremental Classifier and Representation Learning}.
\newblock In {\em The IEEE Conference on Computer Vision and Pattern
  Recognition (CVPR)}, Honolulu, Hawaii, 2017.

\bibitem{Rusu16progressive}
A.~A. {Rusu}, N.~C. {Rabinowitz}, G.~{Desjardins}, H.~{Soyer},
  J.~{Kirkpatrick}, K.~{Kavukcuoglu}, R.~{Pascanu}, and R.~{Hadsell}.
\newblock {Progressive Neural Networks}.
\newblock {\em ArXiv e-prints}, June 2016.

\bibitem{Shin17}
Hanul Shin, Jung~Kwon Lee, Jaehong Kim, and Jiwon Kim.
\newblock Continual learning with deep generative replay.
\newblock In {\em Advances in Neural Information Processing Systems}, pages
  2990--2999, 2017.

\bibitem{wu2018memory}
Chenshen Wu, Luis Herranz, Xialei Liu, yaxing wang, Joost van~de Weijer, and
  Bogdan Raducanu.
\newblock Memory replay gans: Learning to generate new categories without
  forgetting.
\newblock In S.~Bengio, H.~Wallach, H.~Larochelle, K.~Grauman, N.~Cesa-Bianchi,
  and R.~Garnett, editors, {\em Advances in Neural Information Processing
  Systems 31}, pages 5962--5972. Curran Associates, Inc., 2018.

\bibitem{wu2019large}
Yue Wu, Yinpeng Chen, Lijuan Wang, Yuancheng Ye, Zicheng Liu, Yandong Guo, and
  Yun Fu.
\newblock Large scale incremental learning.
\newblock In {\em Proceedings of the IEEE/CVF Conference on Computer Vision and
  Pattern Recognition}, pages 374--382, 2019.

\bibitem{Zenke17}
Friedeman {Zenke}, Ben {Poole}, and Surya {Ganguli}.
\newblock Continual learning through synaptic intelligence.
\newblock In Doina Precup and Yee~Whye Teh, editors, {\em Proceedings of the
  34th International Conference on Machine Learning}, volume~70 of {\em
  Proceedings of Machine Learning Research}, pages 3987--3995, International
  Convention Centre, Sydney, Australia, 06--11 Aug 2017. PMLR.

\end{thebibliography}

\end{document}